\documentclass[runningheads]{llncs}
\usepackage[T1]{fontenc}
\usepackage{graphicx}
\usepackage{amsmath}
\usepackage[table]{xcolor}
\usepackage{colortbl}
\definecolor{lightgray}{gray}{0.94}
\usepackage{hyperref}
\usepackage{cite}
\usepackage{color}

\begin{document}
%
\title{Predicting Stroke through Retinal Graphs and Multimodal Self-supervised Learning}
\titlerunning{Retinal Graphs and Multimodal Self-supervised Learning}
%
\author{Yuqing Huang\inst{1}
\and
Bastian Wittmann\inst{3}
\and
Olga Demler\inst{1,4}
\and
Bjoern Menze\inst{3}
\and
Neda Davoudi\inst{1,2,3}\thanks{Corresponding author}
}

%
\authorrunning{Y. Huang, B. Wittmann, O. Demler, B. Menze, \& N. Davoudi}
%
\institute{Affiliation}
\institute{ETH Zürich, Zürich, Switzerland \\
\and
ETH AI Center, ETH Zürich, Zürich, Switzerland\\
\email{neda.davoudi@ai.ethz.ch} 
\and
Department of Quantitative Biomedicine, University of Zurich, Zürich, Switzerland
\and
Harvard Medical School, Boston, Massachusetts, USA}

%
\maketitle              
\begin{abstract}
Early identification of stroke is crucial for intervention, requiring reliable models. We proposed an efficient retinal image representation together with clinical information to capture a comprehensive overview of cardiovascular health, leveraging large multimodal datasets for new medical insights. Our approach is one of the first contrastive frameworks that integrates graph and tabular data, using vessel graphs derived from retinal images for efficient representation. This method, combined with multimodal contrastive learning, significantly enhances stroke prediction accuracy by integrating data from multiple sources and using contrastive learning for transfer learning. The self-supervised learning techniques employed allow the model to learn effectively from unlabeled data, reducing the dependency on large annotated datasets. Our framework showed an AUROC improvement of 3.78\% from supervised to self-supervised approaches. Additionally, the graph-level representation approach achieved superior performance to image encoders while significantly reducing pre-training and fine-tuning runtimes. These findings indicate that retinal images are a cost-effective method for improving cardiovascular disease predictions and pave the way for future research into retinal and cerebral vessel connections and the use of graph-based retinal vessel representations.

\end{abstract}
\section{Introduction}
Cardiovascular diseases (CVDs) are the leading cause of death worldwide
~\cite{lindstrom2022global}, which makes proactive monitoring of risk factors a critical task in medical research. 
One of the major subclasses of CVDs is stroke, 
a medical condition in which poor blood flow to the brain causes cell death and makes the brain stop functioning properly. 
Deep learning (DL) contributes to stroke treatment by detecting infarcts or hemorrhages, segmenting images, identifying large vessel occlusions, early detection, and providing prognostic insights~\cite{wardlaw2022accuracy,tan2023early}. 


Recent studies suggest that retinal microvascular changes reflect cerebral small vessel disease (CSVD)~\cite{ji2024predicting} and cardiovascular disease (CVD)~\cite{poplin2018prediction,tseng2023validation}. The RETFound model was developed to generalize disease detection from retinal images~\cite{zhou2023foundation}. Retinal microvascular biomarkers may indicate systemic cardiovascular diseases like hypertension, atherosclerosis, and heart failure~\cite{kellner2024eye}. Retinal vessel thickness is linked to intracranial hypertension~\cite{kwapong2023retinal}, while retinal vessel geometry is associated with MRI markers and CVD phenotypes in elderly participants~\cite{arnould2023retinal}. Recent works focus on early diagnosis and risk management of CVD using retinal images with DL algorithms~\cite{sheela2024revolutionizing}. 

To leverage the huge amount of unannotated data and assess the risk of CVDs more efficiently, we propose a contrastive learning framework on fundus photographs (FPs) and clinical information in tabular format. Our framework is developed upon a general contrastive learning model using tabular and imaging data~\cite{BestofBothWorlds}. In addition, we provide three different modules to extract embeddings from fundus photographs. The first module, based on ResNet~\cite{He2015DeepRL}, takes the raw fundus photographs as input. The second module transforms retinal images into probabilistic vessel masks, which are then processed by a ResNet model. Finally, the third module constructs a vessel graph representation from the fundus photograph and uses Graph Neural Networks (GNNs) to learn the feature embedding. This is the first time, to the best of our knowledge, that retinal images are represented as vessel graphs and further utilized in a contrastive learning framework. We show the effectiveness of the contrastive pre-training by comparing it against other baseline approaches for stroke prediction. 
Our code is publicly available at \url{https://github.com/yuqinghuang01/MMCL-Tabular-Fundus}.

\subsection{Multimodal learning}
Integrating patient data from diverse sources in real-time facilitates more effective prevention and treatment strategies. 
Previous work showed that the fused clinical and imaging models outperformed models that included only one modality~\cite{liu2023functional}. 
DL models, leveraging extensive data, are recognized as valuable tools for 
enhanced diagnosis and multimodal prognostication, particularly in the rapidly advancing field of stroke imaging ~\cite{cui2022deep}.
Current models are limited by being trained on a single modality and not translating between different modalities. To fully capture the complexity of human biology, it's necessary to go beyond traditional expert-curated features and include other important data types that doctors also rely on~\cite{gong2022supervised}. ML models can leverage the complementary information present in different modalities to develop a joint characterization of physiological states and further enhance their effectiveness. 
Lee et al. developed an AI model to identify CVD using multimodal data, including clinical risk factors and fundus photographs via supervised learning~\cite{lee2023multimodal}. 

\subsection{Self-supervised Learning}
Recent models excel in biomedical tasks, but overfitting risks persist due to limited annotated medical datasets for supervised learning. Results of generative models such as Autoencoders (AEs)~\cite{alain2014regularized} on multimodal clinical measurements show that they perform well on aligning the embeddings from diverse modalities and constructing a holistic representation for characterizing physiological states~\cite{radhakrishnan2023cross}. 
AEs are employed to learn cross-modal representations from large multimodal datasets. The UK Biobank~\cite{sudlow2015uk} serves as an excellent resource for learning clinically relevant representations. Hager et al. attempted to combine images and tabular data for pre-training of representation by optimizing a CLIP loss and predicting myocardial infarction (MI)~\cite{BestofBothWorlds}. 
Diaz et al. estimated risks of future events of MI using diagnostic features from retinal images that may be undiscovered by human experts~\cite{diaz2022predicting}. The study estimates cardiac indices, such as left ventricular mass (LVM) and left ventricular end-diastolic volume (LVEDV), and predicts future MI events with multi-channel variational autoencoder. The learned latent space is used to train the ResNet model with cardiac MRI reconstructed from the retinal images plus the demographic data to estimate LVM and LVEDV. Finally, they predict the MI risk using logistic regression.

\subsection{Graph Representation of Vessels}
Graphs efficiently represent vessel structures, enabling GNNs to model their topological and geometric properties for accurate segmentation. Unlike Convolutional Neural Networks (CNNs), which work on a regular image grid, GNNs excel in capturing complex vascular connectivity globally.
Shin et al. proposed Vessel Graph Network (VGN) that combines a GNN into a comprehensive CNN architecture for vessel segmentation~\cite{SHIN2019101556}. 
They showed that VGN performs better at segmenting thinner vessels and suppressing false positives by considering the graphical vessel structure. Mishra et al. exploited graph convolutional networks (GCN) such that the model can benefit from vessel topological features for retinal artery/vein classification~\cite{mishra2021vtg}. Recent research converted vessels into graph representations to efficiently enhance the accuracy and connectivity of vessel segmentation ~\cite{li2022graph,yu2022vessel}. 
Drees et al. provide an open-source tool for scalable, robust graph and feature extraction from vessel segmentations~\cite{drees2021scalable}. The method generates node features including position and degree as well as edge features such as length, curveness, and volume. This tool, among others~\cite{bumgarner2022vesselvio}, is a versatile solution for generating graph-level representations for segmented vasculatures. 

\section{Contrastive Learning Frameworks}
\begin{figure}[h]
\includegraphics[width=\textwidth]{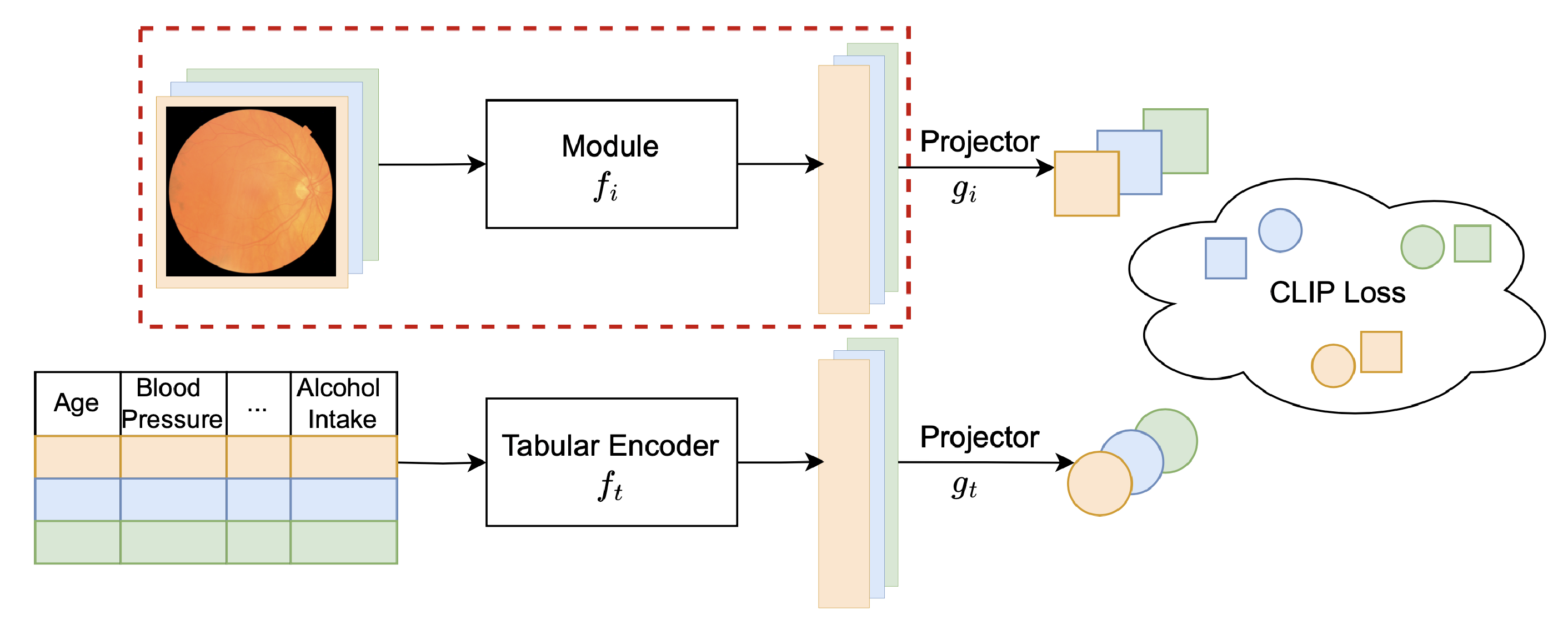}
\caption{The overall contrastive learning framework with fundus photographs and tabular data. The Module in the dotted box will be further specified in the subsections depending on the strategy used.} \label{fig_overall}
\end{figure}
Our proposed framework (Fig.~\ref{fig_overall}) is based on the backbone of multimodal contrastive learning~\cite{BestofBothWorlds} where the data sample pair $x=(x_i, x_t)$ consists of imaging data $x_i$ and tabular data $x_t$. The encoder module for imaging data is denoted as $f_i$ and that for tabular data as $f_t$. In the following subsections, we discuss three encoder modules proposed for the imaging modality. The modules are introduced in the order of increasing preprocessing steps to condense the information in fundus photographs. The projectors for imaging and tabular data are $g_i$ and $g_t$, respectively. The projected embeddings for the two data modalities are
\[
x'_i = g_i(f_i(x_i)) \quad \text{and} \quad x'_t = g_t(f_t(x_t))
\]

As $x'_i$ and $x'_t$ are projections in the same latent space, a CLIP loss~\cite{pmlr-v139-radford21a} is employed to pull projections from the same sample closer and push away projections from different samples. Considering a batch $\mathcal{B}$ of data pairs, the total loss is therefore a weighted sum of imaging modality loss $\ell_{i,t}$ and tabular modality loss $\ell_{t,i}$. Note that $\ell_{t,i}$ is defined symmetrically to $\ell_{i,t}$, and $\tau,\lambda$ are both hyperparameters.
\begin{equation}
\mathcal{L} = \lambda \ell_{i,t} + (1-\lambda) \ell_{t,i}
\end{equation}
\begin{equation}
    \ell_{i,t} = -\sum_{x\in \mathcal{B}} \log {\frac{\exp{(\cos{(x'_i, x'_t)/\tau)}}}{\sum_{y\in \mathcal{B}\setminus\{x\}} \exp{(\cos{(x'_i,y'_t) / \tau)}}}}
\end{equation}

\subsection{Raw Fundus Photographs}
\begin{figure}[h]
\centering
\includegraphics[width=0.5\textwidth]{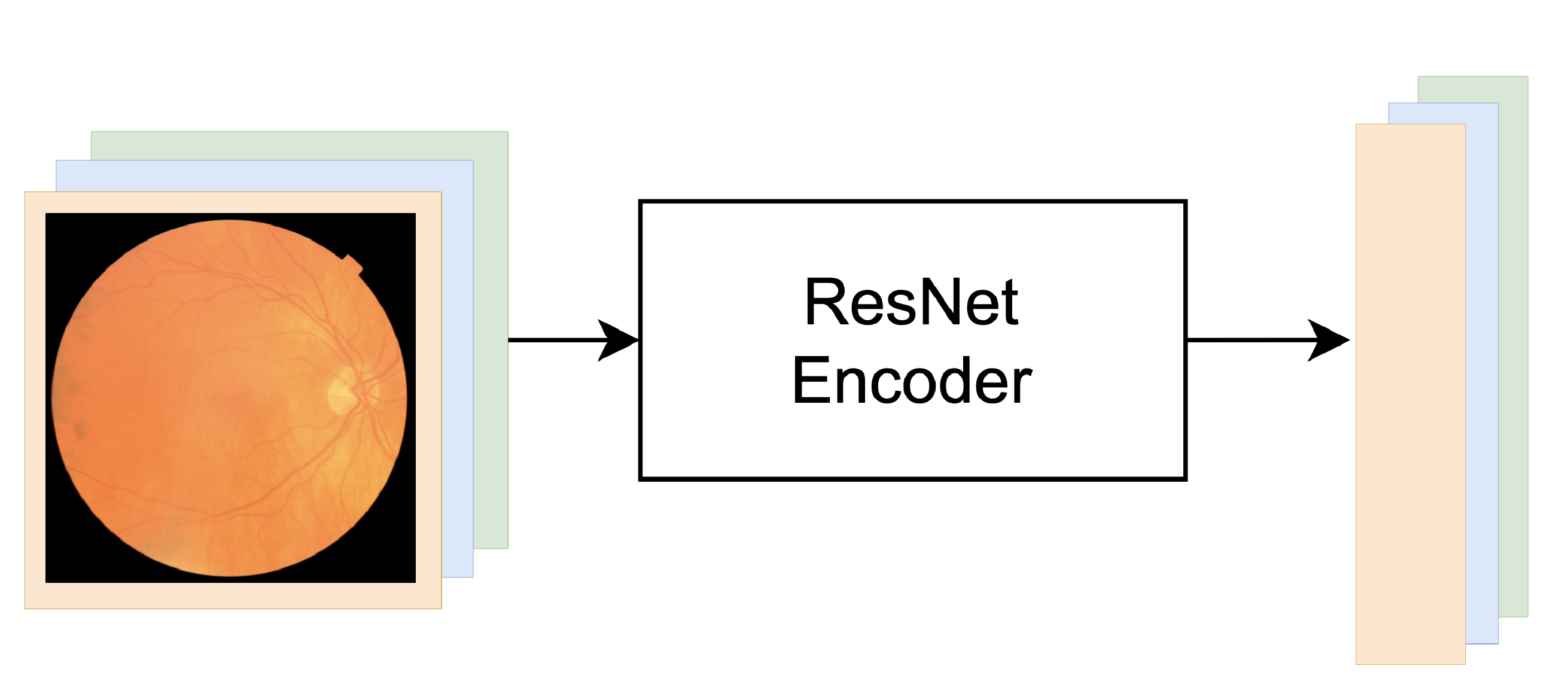}
\caption{Module that directly learns from raw fundus images.} \label{fig_image}
\end{figure}

Firstly, we use ResNet50~\cite{He2015DeepRL} as the imaging encoder. As illustrated in Fig.~\ref{fig_image}, the encoder directly takes raw fundus images and generates embeddings of fixed size. We name this contrastive learning method Multimodal-CL-raw.

\subsection{Retinal Vessel Probability Mask}
\begin{figure}[h]
\centering
\includegraphics[width=0.8\textwidth]{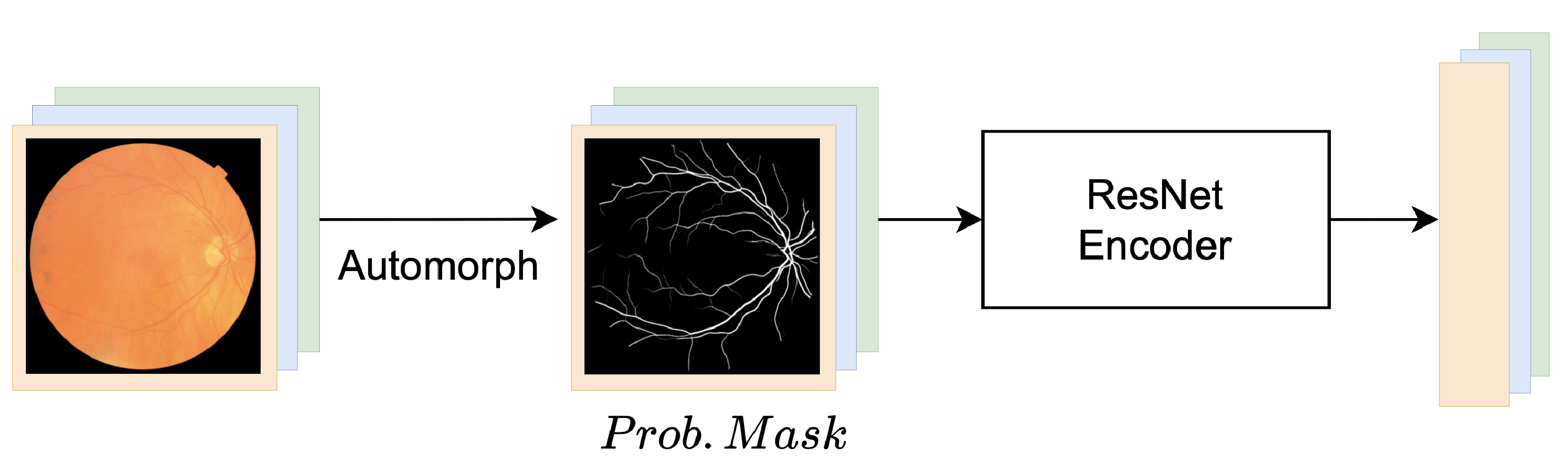}
\caption{Module that extracts probabilistic masks from fundus images.} \label{fig_prob}
\end{figure}

We propose a two-step processing module (Fig.~\ref{fig_prob}). The first part consists of generating vessel probability masks from raw retinal images using the Automorph pipeline~\cite{zhou2022automorph}. The second part is a ResNet-based encoder producing embedding vectors. This contrastive learning method is referred to as Multimodal-CL-prob.

\subsection{Retinal Vessel Graph}
\begin{figure}[h]
\includegraphics[width=\textwidth]{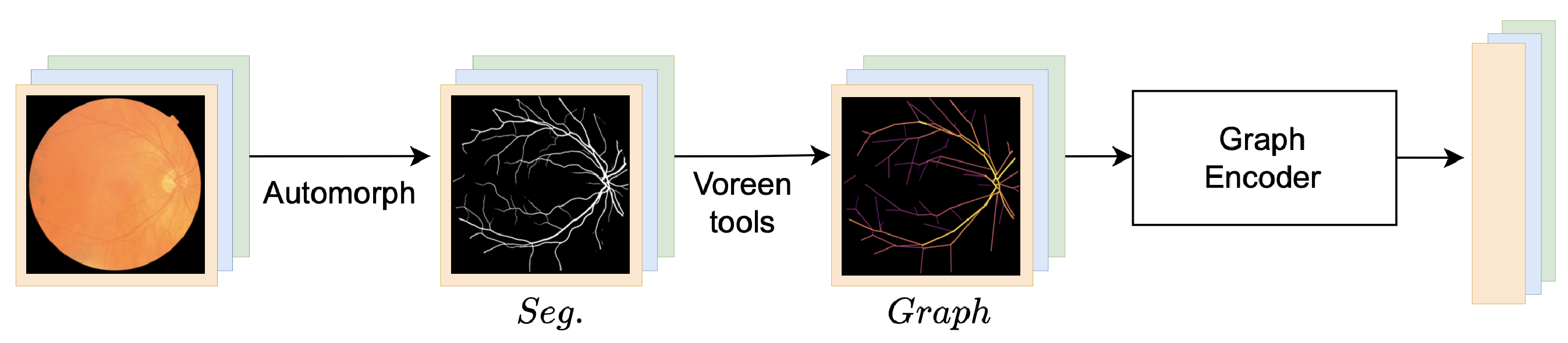}
\caption{Module that turns fundus images into vessel segmentations and further into graph representations. A graph encoder is used for feature extraction from graphs.} \label{fig_graph}
\end{figure}

As illustrated in Fig.~\ref{fig_graph}, the third module exploits graph-level representation for retinal vessels. Pre-processing steps generate binary vessel segmentations using Automorph~\cite{zhou2022automorph} and extract vessel graphs and features using Voreen tools~\cite{voreen}. Finally, the graphs are processed by a graph encoder consisting of graph attention layers~\cite{velickovic2018graph} and a global pooling layer to produce the feature embedding. We name this contrastive learning method Multimodal-CL-graph.

\section{Experiments and Results}
\subsection{Dataset}
Our analyses were performed on UK Biobank, which is a prospective cohort study of over 500,000 individuals' data from across the United Kingdom \cite{sudlow2015uk}. 
This massive database includes patients' information as well as fundus images. 
To demonstrate the effectiveness of the contrastive pre-training, our goal was to predict future risk of stroke. Therefore, participants who have stroke diagnosed after data collection were labeled as positive instances in the dataset.

For the tabular data, 49 features are selected from the following categories: demographics, biomarkers, comorbidities, lifestyles, and medications. The binary and categorical data fields were one-hot encoded, while the continuous data fields were normalized to a mean value of 0 and standard deviation of 1. Missing values in the tabular data were imputed using an iterative imputer. For the imaging data, the raw fundus photographs were cropped at the center to remove the unnecessary background and min-max normalized to the range between 0 and 1. Transformations such as random flip, random rotation, and color jittering were applied to the images in the contrastive training. Finally, the images were resized to $128\times 128$ using bicubic interpolation. No data augmentations were applied for the graph modality as we would like to preserve the vessel structures.

The total size of the dataset is 82569. 20\% was used for generating a test set, while the rest are further split with an 80-20 ratio into a training set and a validation set.
After enriching our embedding with multimodal contrastive learning, we fine-tune the encoders in a supervised fashion by training on a balanced dataset comprises all patients with diagnosed disease and randomly chosen individuals from the healthy cohort.

\subsection{Experimental Setup}
The imaging encoder is based on ResNet50~\cite{He2015DeepRL}, while the tabular encoder, tabular projector, and imaging projector are multilayer perceptions (MLPs) with one hidden layer. The tabular encoder is an MLP with a hidden layer of size 1024 and an output layer of size 1024. In the method Multimodal-CL-graph, the graph encoder consists of three graph attentional layers, using both node and edge features from the vessel graphs.
The three layers were defined to have 4, 4, and 2 attention heads and 10, 50, and 256 output channels, respectively. Hence, the graph encoder generates embeddings of dimension 512. Adam optimizer was used with a batch size of 64 in all experiments.
We report area under the receiver operating characteristic curve (AUROC) on the test set, as used in previous works on CVD prediction~\cite{radhakrishnan2023cross}~\cite{BestofBothWorlds}~\cite{lee2023multimodal}~\cite{diaz2022predicting}. 

For downstream task, the three contrastive learning methods used both tabular features and images as inputs. Our goal was to investigate whether contrastive pre-training on all available data pairs could be beneficial for stroke prediction. To evaluate the performance of our methods Multimodal-CL-raw, Multimodal-CL-prob, and Multimodal-CL-graph, we compared them against three fully supervised baselines on stroke prediction. We also tested the XGBoost on tabular data, which gives much lower AUROC 
comparing to neural networks (NNs). Therefore, we focused on NN-based methods in the experiments for consistency. For these supervised methods, we used the same 80-20 train-test split ratio. Furthermore, the train and validation sets are balanced so that the supervised baselines are in fair comparison with the self-supervised methods.
\paragraph{\textbf{Tabular-NN}}
We defined a neural network (NN) with one hidden layer of size 256 and an output of size 2. The model was developed using tabular data only.
\paragraph{\textbf{Imaging-NN}}
Imaging-NN is composed of an encoder network and a classifier. 
For the encoder network, we used the architecture and pre-trained weights taken from a foundation model, RETFound~\cite{zhou2023foundation}. The RETFound model is a masked autoencoder trained on millions of color fundus photographs. Specifically, the encoder uses a Vision Transformer (ViT)~\cite{dosovitskiy2021an} with 24 Transformer blocks and creates an embedding vector of size 1024. The classifier takes 1024-dimension inputs, features a width-512 hidden layer, and outputs a prediction of size 2. The weights of the pre-trained RETFound encoder were frozen throughout the training and testing phases.
\paragraph{\textbf{Multimodal-NN}}
Multimodal-NN takes both the tabular and imaging modality. For the tabular data, the model learns feature embeddings of size 256 using an MLP. Embeddings for the imaging modality were learned the same way as in Imaging-NN method, using the pre-trained RETFound encoder. In addition, we projected the size-1024 imaging embedding vectors to dimension 256 and concatenated them with the tabular embeddings to make the final prediction using a fully connected layer.

\subsection{Stroke Prediction}


Table~\ref{tab_test} shows the evaluation metrics reported on the test set. 
The results of Tabular-NN, Imaging-NN, and Multimodal-NN indicate that tabular data plays an important role in stroke prediction. 
Compared to using tabular modality only, concatenating the learned features from tabular and imaging modality in a fully supervised setting seemed not sufficient for improving the AUROC.
With contrastive pre-training, our proposed approaches outperformed all supervised baselines, with Multimodal-CL-graph achieving the best AUROC 71.92\%. Multimodal-CL-prob also demonstrated better performance than Multimodal-CL-raw on the test set. 

\begin{figure}[h]
\centering
\includegraphics[width=0.6\textwidth]{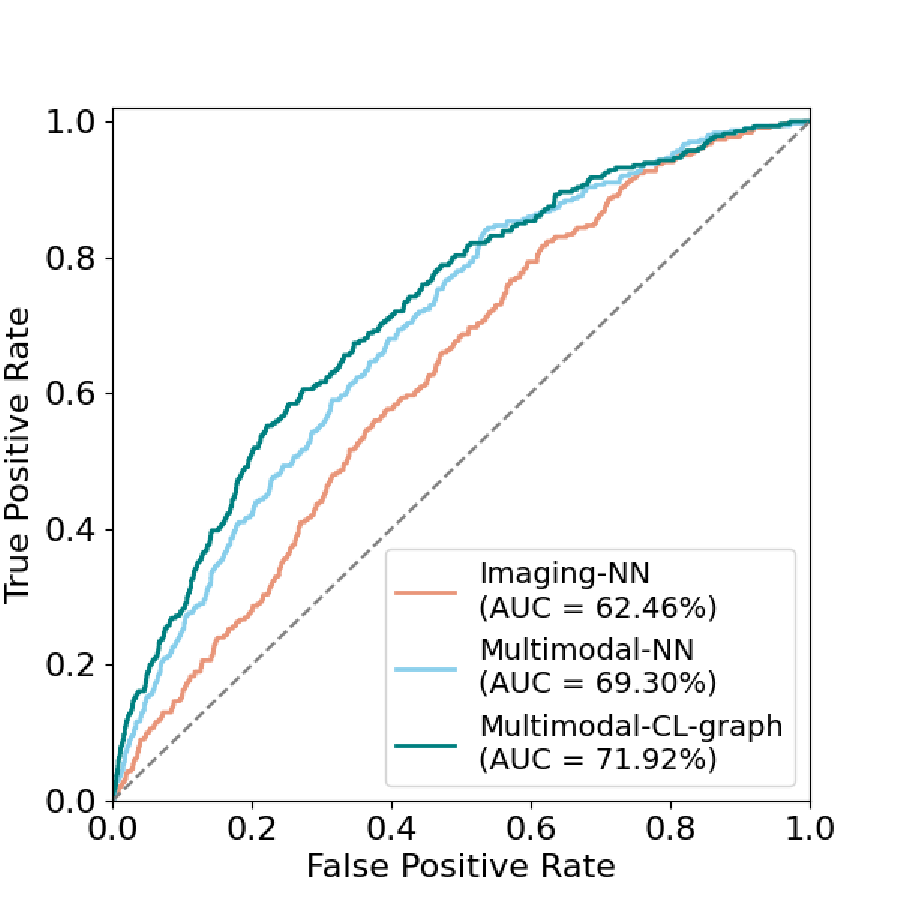}
\caption{Comparison of test-time ROC curves for Imaging-NN, Multimodal-NN, and Multimodal-CL-graph methods.} \label{fig_test_rocs}
\end{figure}

Furthermore, we showed a visualization (Fig.~\ref{fig_test_rocs}) of the receiver operating characteristic (ROC) curves of three methods: Imaging-NN, Multimodal-NN, and Multimodal-CL-graph. Out of the two fully supervised methods, multimodal training with tabular and imaging data outperformed unimodal training with imaging data, which shows that fundus images are best used as an additional modality rather than the only data modality for prediction. Compared to the multimodal supervised baseline, making use of unlabelled paired data in self-supervised pre-training gives further improvement in predicting future risk of stroke.


\begin{table}[h]
\caption{A summary table of evaluation metrics for different configurations of supervised and self-supervised methods. Our methods are highlighted in gray.}\label{tab_test}
\centering
\begin{tabular}{|>{\centering\arraybackslash}p{3.5cm}|>{\centering\arraybackslash}p{2.4cm}|}
\hline
\textbf{Method} &  \textbf{AUROC(\%)} \\
\hline
Tabular-NN & 70.02 \\
Imaging-NN & 62.46 \\
Multimodal-NN & 69.30 \\
\rowcolor{lightgray}
Multimodal-CL-raw & 71.54 \\
\rowcolor{lightgray}
Multimodal-CL-prob & 71.73 \\
\rowcolor{lightgray}
Multimodal-CL-graph & \textbf{71.92} \\
\hline
\end{tabular}
\end{table}

\subsection{Efficiency Comparison}
In table~\ref{tab3}, we compare the sizes of the networks and the runtimes per epoch when using image encoder versus when using graph encoder in the contrastive learning framework. Vessel graphs with node and edge features are compact representations of the vessel structure in retinal images. The number of trainable parameters in the contrastive pre-training with graph encoder is 3.4 M, which is about 12\% of those needed for contrastive pre-training with image encoder. In terms of training time, pre-training takes about 840 seconds per epoch with ResNet50 encoder and about 90 seconds per epoch with graph attention network (GAT) encoder. Similarly, during fine-tuning, Multimodal-CL-graph requires significantly fewer trainable parameters and thus less training time per epoch compared to Multimodal-CL-image or Multimodal-CL-prob while outperforming the test-time AUROCs of the latter two.
\begin{table}[h]
\caption{Comparison of computational efficiency for image and graph encoders by trainable parameters and runtime per epoch during pre-training and fine-tuning.
}\label{tab3}
\centering
\begin{tabular}{|>{\centering\arraybackslash}p{3.4cm}|>{\centering\arraybackslash}p{1.8cm}|>{\centering\arraybackslash}p{1.8cm}|>{\centering\arraybackslash}p{1.8cm}|>{\centering\arraybackslash}p{1.8cm}|}
\hline
& \multicolumn{2}{c|}{\textbf{Pre-training}} & \multicolumn{2}{c|}{\textbf{Fine-tuning} }\\
\hline
 &  Parameters & Per epoch &  Parameters & Per epoch\\
\hline
Multimodal-CL-raw/ Multimodal-CL-prob & 28.2 M & 840 s & 24.8 M & 13 s\\
\hline
Multimodal-CL-graph & 3.4 M & 90 s & 1.5 M & 2 s\\
\hline
\end{tabular}
\end{table}

\section{Discussion and Conclusion}
In this work, we presented self-supervised methods leveraging retinal images and medical information in tabular format to improve stroke prediction. Moreover, we proposed the first contrastive framework that combines graph and tabular data. In particular, vessel graphs were extracted from retinal images for a more efficient representation and were paired with tabular features in multimodal contrastive learning. 
The contrastive pre-training step allowed us to use all available paired tabular-fundus data, even in the absence of labels, by jointly learning a latent space. 
We improved the AUROC from supervised to self-supervised methods by 3.78\%.
In addition, compared to the methods using image encoders, the method using graph-level representations achieves superior performance while being more efficient with regard to number of parameters and around ten times faster in terms of pre-training and fine-tuning runtimes.

Here are a few directions for improving the study. First, our experiments focus on the downstream task of stroke prediction. Future work can also examine the effectiveness on other cardiovascular diseases as downstream classification tasks such as myocardial infraction prediction. 
Second, for the fine-tuning datasets, we randomly sampled the healthy people with the same number of people having stroke. 
Further ablation studies can be conducted to investigate how the ratio between the positive and negative cases in the fine-tuning set would affect the performance of downstream tasks.
Third, the graph encoder used in our experiments consists of graph attention layers and a global pooling layer to extract the final embedding vector. Future studies can explore the use of other graph encoder architectures and test their effectiveness. For example, Ying et al.~\cite{ying2018hierarchical} proposed a differentiable graph pooling module that generates hierarchical representations of graphs. This method offers an alternative solution to global pooling and can be adapted to aggregate features in vessel graphs in a hierarchical manner.

In conclusion, we proposed novel approachs to integrate the fundus photographs in a contrastive learning framework. Our findings suggest that retinal images could be a cost-effective modality for improving stroke predictions. We believe these results are clinically meaningful and would motivate future research exploring the relation between retinal and cerebral vessels as well as using graphs as a compact representation of retinal vessels.


\begin{credits}
\subsubsection{\ackname} This research used UK Biobank under application number 81959. We thank Prof. Samia Mora from Harvard Medical School for her invaluable medical insight and guidance.
This project was supported by the grant \#2023-N-306 of the 1st Joint Call of the Swiss Data Science Center (SDSC) and the Strategic Focus Area “Personalized Health and Related Technologies (PHRT)” of the ETH Domain (Swiss Federal Institutes of Technology). N.D. is partially supported by the ETH AI Center postdoctoral fellowship. B.M., B.W., and N.D. acknowledge support of the Helmut-Horten-Foundation.

\subsubsection{\discintname}
The authors have no competing interests to declare that are relevant to the content of this article. 
\end{credits}
%
%
%
\bibliographystyle{splncs04}
\bibliography{library}
%

%
%
%

\end{document}